# How do some Bayesian Network machine learned graphs compare to causal knowledge?

Anthony C. Constantinou, Norman Fenton, and Martin Neil

Nov 2018

**Abstract**— The graph of a Bayesian Network (BN) can be machine learned, determined by causal knowledge, or a combination of both. In disciplines like bioinformatics, applying BN structure learning algorithms can reveal new insights that would otherwise remain unknown. However, these algorithms are less effective when the input data are limited in terms of sample size, which is often the case when working with real data. This paper focuses on purely machine learned and purely knowledge-based BNs and investigates their differences in terms of graphical structure and how well the implied statistical models explain the data. The tests are based on four previous case studies whose BN structure was determined by domain knowledge. Using various metrics, we compare the knowledge-based graphs to the machine learned graphs generated from various algorithms implemented in TETRAD spanning all three classes of learning. The results show that, while the algorithms produce graphs with much higher model selection score, the knowledge-based graphs are more accurate predictors of variables of interest. Maximising score fitting is ineffective in the presence of limited sample size because the fitting becomes increasingly distorted with limited data, guiding algorithms towards graphical patterns that share higher fitting scores and yet deviate considerably from the true graph. This highlights the value of causal knowledge in these cases, as well as the need for more appropriate fitting scores suitable for limited data. Lastly, the experiments also provide new evidence that support the notion that results from simulated data tell us little about actual real-world performance.

——————————— ◆ ———————————

## 1 Introduction

STATISTICAL regression and classical machine learning methods ignore assumptions of causation, or the direction of influence between factors of interest. As a result, such methods generally lead to models similar to Model *A* in Fig 1 (the 'associative' model), which simply states that each factor is somehow associated to each other and hence, predictive of each other. On the other hand, Model *B* (the 'causal' model) is a Directed Acyclic Graph (DAG) and captures the same features under causal or influential assumptions.

Unlike association, causal assumptions make claims about the effect of interventions. Specifically, unlike the associative model in Fig 1, the causal model tells us that an intervention on *Yellow teeth* will have no effect on *Smoking* nor on *Lung cancer*, whereas an intervention on *Smoking* will have an effect on both *Yellow teeth* and *Lung cancer*. In the associative model, the association between Yellow teeth and Lung cancer comes via the common cause *Smoking*. In the causal model, a link between *Yellow teeth* and

*Lung cancer* would be incorrect, and such incorrect causal connections can be discovered and eliminated in causal models through knowledge, through algorithms that test for conditional independence tests, or a combination of the two.

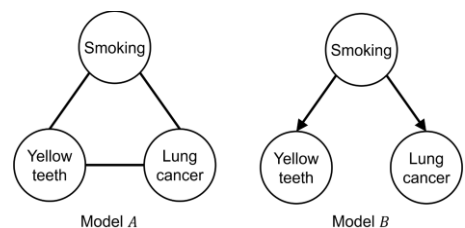

**Fig 1.** Two hypothetical models under assumptions of association (Model *A*) and causation (Model *B*).

Much of the research in discovering relationships between information is based on methods which focus on maximising the predictive accuracy of a targeted variable *X* from a set of observed predictors *Y*. However, the best predictors of *X* are often not the causes of *X*, hence the motto "*association does not imply causation*". Although the distinction between association and causation is nowadays better understood, what has changed over the last few decades is mostly the way the results are stated rather than the way they are generated. While scientific research is heavily driven by interest in discovering, assessing, and modelling real-world cause-and-effect relationships as guides for

- *A.C. Constantinou is with the School of Electronic Engineering and Computer Science, Queen Mary University of London, London, UK, E1 4NS. E-mail: a.constantinou@qmul.ac.uk.*
- *N. Fenton is with the School of Electronic Engineering and Computer Science, Queen Mary University of London, London, UK, E1 4NS. E-mail: n.fenton@qmul.ac.uk.*
- *M. Neil is with is with the School of Electronic Engineering and Computer Science, Queen Mary University of London, London, UK, E1 4NS. E-mail: m.neil@qmul.ac.uk*

action and decision-making, most scientific conclusions continue to be based on results derived from outputs of associative models and hence, often fail to accurately answer the important questions of intervention which require models under causal assumptions.

A Bayesian Network (BN), which is a type of a probabilistic graphical model [1] introduced by Pearl [2], [3] is a DAG that offers a framework for modelling relationships between information under causal or influential assumptions, often referred to as Causal Bayesian Networks (CBNs). As a result, BNs enable us to construct models similar to the causal model in Fig 1, where the nodes represent uncertain variables and the arcs represent the direction of influence. BNs are also recognised as the most appropriate method to model uncertainty in situations where the model could benefit from both data and available knowledge. This is especially true in *Bayesian Decision Networks* (BDNs), also known as *Influence Diagrams*, which represent an extended version of BNs suitable for optimised decision-making. BDNs require that we specify the decision options available to the decision makers, along with the utilities we seek to minimise or maximise [4], [5]. However, in this paper, we shall focus on the standard representation of BNs.

Pearl's and Mackenzie's recent book [6] has brought great attention to the importance and need for causal models like BNs as the basis for achieving true AI. The question of how to most effectively develop the necessary BNs to solve real-world problems is therefore a particularly current concern. There are two main steps in constructing BNs. First, we establish the causal or influential (hereafter referred to as *causal*[1]) graph of the model, also known as BN 'structure' (hereafter referred to as *graph*), prior to parameterising the Conditional Probability Tables (CPTs) that specify the strength and shape of influence between parents and children.

In this paper, we are interested in the construction of the causal graphs. While BNs provide an effective graphical representation of a problem, and can be used for multiple types of complex inference [7], an enduring challenge is to construct an accurate causal graph. In general, there are three ways to construct a BN graph: a) knowledge, b) causal discovery or BN structure learning algorithms, or c) a combination of both, such as introducing knowledge-based constraints in terms of what can and cannot be discovered by an algorithm. Further, there are primarily three machine learning approaches to learn BN graphs [8]:

i. **Constraint-based:** These algorithms aim to establish links between variables under the assumption that the arcs represent causal relationships. They primarily rely on conditional independence checks between variables in sets of triples, a process inherited from the *Inductive Causation* (IC) algorithm [9]. The *Peter and Clark* (PC) algorithm has had major impact in this area of research due to its simplicity and learning strategies [10], [11]. These algorithms tend to be evaluated solely on their ability to recover a ground truth network from data, although some algorithms allow the option to incorporate knowledge into the structure learning process; e.g., variable $B$ occurs after $A$ and hence $B$ cannot influence $A$. Different knowledge-based constraints can be found in [12], [13].

ii. **Score-based:** These algorithms search for different structures and score them using an objective function that measures how well the fitting distributions agree with the empirical distributions, often relative to model complexity. A large number of algorithms fall within this area of research [14], [15]. Examples include the K2 algorithm [16], Sparse Candidate Algorithm [17], the Optimal Reinsertion algorithm [18], and *Greedy Equivalence Search* (GES) [19]. As in (*i*) above, some of these algorithms also allow knowledge to be incorporated into the process of structure learning.

iii. **Hybrid algorithms:** These combine the properties of the two above types of structure learning, thereby performing both constraint-based and score-based learning. Examples include the *max-min hill-climbing* (MMHC) algorithm [20] and L1-Regularisation paths [21].

While each approach has its strengths and weaknesses, it is widely acknowledged that accurate BN structure learning is a very challenging task due to the fundamental problem of inferring causation from observational data which is generally NP-hard [8], [14], [15], [22]. Specifically, the search space of possible graphs that could explain the data is super-exponential to the number of variables; although problems related to learning statistical dependencies from finite data are relaxed, or become irrelevant, when the number of variables is small.

Still, there are some conflicting opinions about what the algorithms can and cannot do [23], [24], [25], [26], [27], [28]. In general, the algorithms demonstrate promising performance when tested with synthetic data (also known as 'simulated' data), which represent hypothetical observations generated from simulation based on predetermined models that are assumed to represent the ground truth. However, the general consensus is that the level of performance observed with synthetic experiments overestimates real-world performance [8], [14], [27], [29], [30], [31]. For more details see Koski and Noble's [8] and Drton and Maathuis' [32] reviews on structure learning.

In this paper we focus on four previously published BN case studies that had the BN graph determined by domain knowledge, and we examine how those knowledge-based graphs compare to the respective machine learned graphs generated using 14 algorithms, including different variants of the same algorithms, implemented in the TETRAD freeware [33]. We describe the methodology in Section 2, we present and discuss the results in Section 3, and we provide our concluding remarks along with directions for future research in Section 4.

---

[1] While BNs and CBNs are assumed to represent dependency and causal networks, when they are applied to real-world problems it is generally because of their ability to represent causal networks.

## 2 METHODOLOGY

The methodology is driven by the aim to assess the differences between purely machine learned and knowledge-based BNs, when both approaches are restricted to the same set of variables used to deduce a BN graph. This assumption is necessary because, as we later show, some of the knowledge-based graphs retrieved include nodes not present in the data. Fig 2 illustrates the overall methodology which we discuss in detail in the subsections that follow. Specifically, we provide details about the case studies in subsection 2.1, the BN structure learning process in subsection 2.2, the algorithms evaluated in subsection 2.3, the evaluation criteria and scoring functions in subsection 2.4, and we discuss the limitations of the methodology in subsection 2.5.

### 2.1 The Case Studies

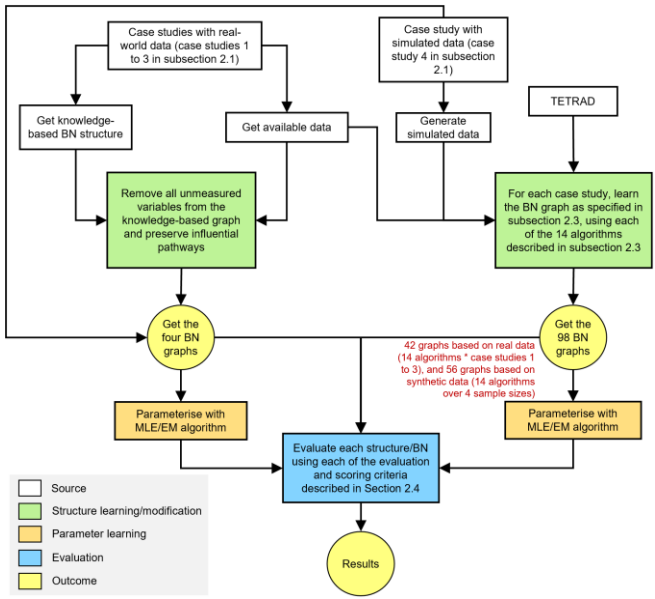

**Fig 2.** The overall methodology.

The four case studies are based on diverse real-world problems, as summarised in Table 1. Fig 2 shows that we have real-world data for case studies 1 to 3, but no data for case study 4. The BN model in case study 4 was developed solely based on knowledge and clearly defined rules[2] in determining both its graph as well as its parameter priors. As a result, case study 4 serves as the simulation-based study in this paper. For case studies 1 to 3, the process involves retrieving the BN graph from the published paper and reducing the graph to the variables available in the dataset (i.e., eliminating latent knowledge-based variables), ensuring that the causal pathways are preserved. This is to ensure that both machine learned and knowledge-based BN graphs are restricted to the same set of measured variables.

TABLE 1.
Key information about the four case studies.

| Case study | Details | Output of interest | Data |
|---|---|---|---|
| 1 (Medicine) | Violence risk analysis in patients discharged from medium secured services [34] developed with psychologists and psychiatrists. | Risk of violence during the first 6 months following discharge from medium secured services. | Variables: 52. States: 2-8. Samples: 386. |
| 2 (Forensic) | Risk assessment and risk management of violent reoffending in released prisoners [35] developed with psychologists and psychiatrists. | Risk of violent reoffending after release from prison, and for up to 5 years following release. | Variables: 56. States: 2-9. Samples: 953. |
| 3 (Sports) | Assesses the evolving performance of football teams and enables us to predict, before a season starts, the total league points a team is expected to accumulate throughout the season [36]. | League points accumulated over the next football season. | Variables: 17. States: 2 to 11. Samples: 300. |
| 4 (Housing market) | Simulation of the UK Buy-To-Let tax reforms, announced in 2015 by the British government, and their impact on the London housing market [37]. | Annual net profit from rental income. | Variables: 27. States: 2 to 7. Samples: 1,000-25,000. |

### 2.2 The BN structure learning process

We made use of the TETRAD v6.2 freeware, which is unique in the suite of exploration and discovery search algorithms and suitable in applying and assessing various BN structure learning methods. The assessment is based on the following key aspects:

i. **Dataset:** We apply BN structure learning to the real-world datasets from case studies 1 to 3, and to the synthetic datasets generated using the BN model from case study 4.

ii. **Data type:** Many algorithms in TETRAD are restricted to either discrete or continuous data. In case studies 1 to 3, almost all of the variables were already discrete. As a result, we focus our assessment on the algorithms that work with discrete data, and discretise the small number of continuous variables such as to ensure that sufficient data points exist for a reasonably well informed prior.

iii. **Structure learning:** TETRAD incorporates 37 structure learning algorithms. However, we were able to assess only 14 out of those 37 algorithms

---

[2] This case study involves a BN that assesses the impact of the major tax reforms introduced by the British government for individual Buy-to-Let (BTL) investors in 2016. The BTL process is based on clearly defined rules and regulating protocols. The BN is entirely based on these manmade, and hence known, rules and protocols.

(refer to Table A.1 for details). While most of the algorithms in TETRAD are constraint-based, the selection of 14 algorithms includes score-based as well as hybrid learning algorithms. Each of the 14 algorithms is briefly introduced in subsection 2.3.

iv. **Parameter learning:** All three real-world datasets incorporate missing values. However, many of the algorithms are restricted to learning from complete datasets. To overcome this limitation, we perform two tests:
   a. we use the original datasets with missing values, and we parameterise the resulting graphs using the Expectation-Maximisation (EM) algorithm which assumes that missing values are missing at random.
   b. we impute missing values with a new state that represents 'missing value' to create a complete dataset, and we parameterise the resulting graphs using Maximum Likelihood Estimation (MLE); this assumes that missing values are not missing at random.

## 2.3 The BN structure learning algorithms

Below we provide a brief introduction to each of the 14 algorithms considered in our experiments, as they have been implemented in TETRAD. Since our tests are restricted to discrete data, we restrict the details about each algorithm to those that relate to discrete data. Some algorithms take different parameters as inputs (discussed below), though due to the scale of the experiments we only consider their default parameters as specified in TETRAD. Detailed technical specification about each of the algorithms is out of the scope of this paper and can be found in the references provided below. In no particular order, the 14 algorithms are:

- **PC:** A constraint-based algorithm that outputs a graph from a set of $d$-separation equivalence classes of DAGs which assume no latent common causes, also known as *pattern* search (see Spirtes et al [11] for more details). The PC algorithm [38] assumes that the input dataset is entirely continuous or entirely discrete. The search process conducts a sequence of marginal and conditional independence tests, starting from a fully connected undirected network, and constructs a pattern graph based on the results of those tests, by eliminating and orientating edges. These tests take a parameter into consideration known as *alpha*, with a default value of 0.01, for rejecting the null hypothesis of independence; i.e., tests with $p$-values greater than 0.01 are judged to be independent. The PC algorithm also takes as input a maximum size of conditioning set (depth); though the default input value considered at -1 allows for unlimited sizes of conditioning sets to be searched.

- **CPC:** A variant of the PC algorithm that aims to increase the precision of colliders (i.e., common effect classes) at the expense of recall, based on the orientation rules in [39]. This makes CPC (Conservative PC) a conservative version of PC [40]. The default input parameters are the same as for PC.

- **PC-Stable:** A variant of the PC algorithm that modifies the search procedure of adjacency discovery such that the adjacency output is not affected by the order of variables as they appear in the data [41], [32]. The default input parameters are the same as for PC.

- **CPC-Stable:** The CPC algorithm with the PC-Stable adjacency modification step [42]. The default input parameters are the same as for PC.

- **PC-Max:** A variant of the PC algorithm, that is based on CPC, and which picks the conditioning set that maximises the $p$-value as an alternative to the voting rule that is used to classify colliders versus non-colliders in CPC [43]. In addition to the default input parameters of the PC algorithm, the PC-Max algorithm also requires that we specify a) whether the heuristic for orientating unshielded colliders should be used to maximise the $p$-value (default is set to 'true'), and b) the maximum depth for the above heuristic in terms of the distance amongst adjacent nodes of $X$ and $Z$ in collider $X \rightarrow Y \leftarrow Z$ (default depth is set to '3').

- **FGES:** The FGES algorithm is a parallelised and an optimised version of the score-based Greedy Equivalence Search (GES) algorithm that was initially developed by Meek [39], and which was later further developed and studied by Chickering [19]. Specifically, the GES algorithm returns the BN that maximises the Bayesian score via heuristic search, and can be applied to discrete or continuous data, or a mixture of the two. FGES (Fast GES) uses the Bayesian Information Criterion (BIC) score [44] to determine the best model. The input parameters of FGES, when working with discrete data, are a) a sample prior added to each CPT (default set to '1'), b) a structure prior coefficient that penalises the number of parents of a variable in any particular CPT (default set to '1'), c) whether the algorithm should assume the faithfulness[3] assumption and remove the edge in $X - Y$ if $X$ and $Y$ are uncorrelated at the initial step in order to speed up the algorithm [45] (default set to 'false'), and d) the maximum number of edges between nodes (default set to '100').

- **IMaGES_BDeu:** A variant of FGES that adjusts the discrete scoring function BDeu [46] to allow for multiple datasets as input. A BDeu score is generated for each dataset and averaged at each step of the algorithm to produce a model over all datasets. Since this study focuses on single datasets, IMaGES_BDeu is expected to produce the results of FGES.

- **CCD:** A variant of the PC algorithm that returns the equivalence class of Partial Ancestral Graphs (PAGs),

---

[3] The *causal faithfulness* condition states that a causal dependency between variables, however it arrives, indicates that there must be a probabilistic dependency.

as opposed to returning the pattern search as in PC [42]. PAGs represent $d$-separation equivalence classes that include models which assume latent common causes and sample selection bias that make them $d$-separation equivalent over a specific set of variables [47].

- **FCI:** A constraint-based algorithm that accounts for the possibility of latent confounders. Similar to the PC algorithm, the Fast Causal Inference (FCI) algorithm performs a series of conditional independence tests to determine which edges to eliminate, starting from a fully connected undirected network. It then proceeds to the orientation phase that uses the stored conditioning sets that had led to the removal of adjacencies at the previous step, to orientate as many of the preserved edges as possible. The FCI algorithm works with discrete data, continuous data, or mixture data of both discrete and continuous values. Two of its input parameters are identical to the PC algorithm; the *alpha* parameter and the depth parameter for the conditioning set. Additionally, the FCI also takes as input the maximum length for any discriminating path during search (default set to unlimited; see Spirtes et al. [11] for details), and whether the FCI rule set should be used (default is 'no') which includes additional orientation rules proposed by Zhang [29].

- **RFCI:** A variant of FCI, the *Really Fast Causal Inference* (RFCI) algorithm, that speeds up search at the expense of a slightly different output that is almost as informative. This is achieved by performing fewer conditional independence tests, while its tests are also conditioned on a smaller number of variables [48]. The default input parameters are the same as for FCI.

- **CFCI:** An experimental algorithm in TETRAD v6.2 that modifies the FCI algorithm in the same way CPC modifies PC.

- **GFCI:** The Greedy Fast Causal Inference (GFCI) algorithm is a hybrid learning algorithm that combines FGES and FCI, to improve both the accuracy and efficiency of FCI. This is achieved by augmenting the initial set on nonadjacencies given by FGES with FCI performing a set of conditional independence tests that lead to the elimination of some additional adjacencies. GFCI then uses some of the orientation rules of FGES to provide an initial orientation of the undirected graph, augmented by the orientation phase of FCI to provide additional orientations [23], [49].

- **FAS:** A variant of the PC algorithm, called Fast Adjacency Search (FAS), that only performs the PC's adjacency search [11]. The default input parameters are the same as for PC.

- **FANG:** An experimental algorithm in TETRAD v6.2 that is being replaced by an algorithm called Fast Adjacency Skewness (FASK). It performs FAS-Stable on the data, which is a variant of the FAS algorithm that modifies FAS in the same way PC-Stable modifies PC, and then performs conditional independence tests to optimise a modified BIC score that accounts for an additional penalty discount $c$ [50]. It shares the same default parameters with PC, plus the penalty discount $c$ (default se to 2).

## 2.4 Evaluation and scoring criteria

All of the resulting BNs are evaluated based on the following criteria:

i. **Degrees of freedom (Df):** In TETRAD, this corresponds to the number of constraints the model entails on the covariance matrix of the probability distribution, and is equal to

$$\frac{m(m-1)}{2} - f$$

where $m$ is the number of measured variables, and $f$ is the number of free parameters, equal to the number of coefficient parameters plus the number of covariance parameters that make up the BN [42].

ii. **Chi squared (Chi²):** How much deviation there is between the estimated model $E$ and data $D$ [42]

$$Chi^2 = \sum \frac{(D-E)^2}{E}$$

iii. **$p$-value ($p$):** The probability, under the assumption the model is true, that we would see at least that much deviation (i.e., Chi² score) between the model and the data [42].

iv. **Bayesian Information Criterion Score (BIC):** A score which rewards likelihood accuracy and penalises models with high parameters that may overfit the data. In TETRAD, the BIC score is calculated as

$$Chi^2 - Df \times log(N)$$

where $N$ is the sample size. The model with the highest BIC score is preferred.

v. **Correctly classified (CC):** The percentage of correctly classified predictions that a BN achieves for a set of variables of interest per case study as specified in the relevant publications (refer to Tables 3, 5, 7, and 9).

vi. **The Area Under the Receiver Operating Characteristic (ROC) Curve (AUC):** While AUC statistics are traditionally used to assess binary outcomes, an AUC statistic of a multinomial distribution can be obtained for each state of the distribution that is measured with respect to all of the other states.

Since TETRAD provides ROC curves for multinomial variables, we report the AUC statistics obtained directly from TETRAD.

vii. **DAG Dissimilarity Metric (DDM):** Quantification of dissimilarities amongst graphs is a major topic in different fields. A number of methods have been proposed for the quantification of network structural dissimilarities [51], [52], but with no particular focus on DAGs. In the area of BN structure learning, the *Structural Hamming Distance* (SHD) is often the preferred choice, and represents the number of arc additions, deletions and reversals needed to convert the learned graph into the ground truth graph. In this paper we use a score that can be viewed as a more complicated version of the SHD score, that which we call *DAG Dissimilarity Metric* (DDM). It assesses the level of dissimilarity between two DAGs. The score ranges from $-\infty$ to 1, where a score of 1 indicates perfect agreement between the two models. The score moves to $-\infty$ the stronger the dissimilarity is between the two models; although this value is bounded by the maximum number of edges a graph entails. Specifically, if we want to compare how dissimilar model $A$ is with respect to model $B$, then

$$DDM = \frac{m + \frac{r}{2} - a - d}{t}$$

where $m$ is the number of arcs matched between the two models, $r$ is the number of arcs from $B$ re-oriented in $A$, $a$ is the number of new arcs added in $A$, $d$ is the number of arcs from $B$ deleted in $A$, and $t$ is total number of arcs in $B$. Since we are interested in analysing how each of the machine learned graphs compare to the knowledge-based graphs, the dissimilarity score DDM is measured with respect to the knowledge-based graph.

## 3 RESULTS AND DISCUSSION

The results are presented per case study in Tables 2 to 9 (two tables per case study). For each of the real-world case studies 1, 2, and 3, one table reports the model statistics under the assumption that missing data are *not*[4] missing at random, and another table reports the predictive accuracy scores for each of the models used to report the model statistics.

The results of case study 4, which are presented in subsection 3.4 and which are based on synthetic data, are based on a sample size of 1,000 observations. Appendix C repeats these results based on sample sizes of 5,000, 10,000, and 25,000 observations.

---

TABLE 2.

Model statistics when trained with the imputed (complete) medical dataset, and parameterised with the MLE algorithm. Top performance scores are underlined.

|           | True model ? | | | | Arc comparison stats | | | | |
|-----------|------|------|---|------|----|----|----|----|--------|
| Algorithm | $Chi^2$ | Df | p | BIC | a | d | r | m | DDM |
| KBG       | 2.6k | 38.4k | 1 | -131k | - | - | - | - | - |
| PC        | 3.8k | 0.5k | 0 | -17.4k | 30 | 68 | 9 | 6 | -1.054 |
| CPC       | 3.8k | 0.5k | 0 | -17.2k | 30 | 68 | 9 | 6 | -1.054 |
| PCS-table | 3.3k | 0.4k | 0 | -17.4k | 20 | 69 | 8 | 6 | -0.952 |
| CPC-Stable | 3.2k | 0.3k | 0 | -17.1k | 20 | 69 | 8 | 6 | -0.952 |
| FGES      | 4.3k | 0.3k | 0 | <u>-16.4k</u> | 30 | 65 | 11 | 7 | -0.994 |
| IMaGES_BDeu | 4.3k | 0.3k | 0 | <u>-16.4k</u> | 30 | 65 | 11 | 7 | -0.994 |
| CCD       | 3.9k | 0.5k | 0 | -17.3k | 30 | 68 | 5 | 10 | -1.030 |
| FANG      | 2.1k | 0.1k | 0 | -17.1k | 19 | 79 | 2 | 2 | -1.145 |
| FCI       | 3.8k | 0.5k | 0 | -17.4k | 30 | 68 | 8 | 7 | -1.048 |
| RFCI      | 3.3k | 0.4k | 0 | -17.4k | 20 | 69 | 7 | 7 | <u>-0.946</u> |
| CFCI      | 3.6k | 0.4k | 0 | -17k | 29 | 68 | 5 | 10 | -1.018 |
| GFCI      | 3.8k | 0.9k | 0 | -18.6k | 30 | 65 | 8 | 10 | -0.976 |
| FAS       | 3.9k | 0.3k | 0 | -16.7k | 29 | 68 | 7 | 8 | -1.030 |

TABLE 3.

Predictive accuracy scores for the primary (P) and secondary (S) variables of interest as specified in the relevant publication [34], based on the models in Table 2. Top performance scores are underlined.

|           | Violent convictions (P) | | General violence (P) | | Anger (S) | | Psychotic illness (S) | | Personalit. disorder (S) | | Violent ideation (S) | |
|-----------|------|------|------|------|------|------|------|------|------|------|------|------|
| Algorithm | AUC | CC | AUC | CC | AUC | CC | AUC | CC | AUC | CC | AUC | CC |
| KBG       | <u>.806</u> | 88.9 | <u>.843</u> | <u>90.4</u> | .936 | <u>86.5</u> | .724 | 63.7 | <u>.838</u> | <u>81.3</u> | <u>.849</u> | <u>74.6</u> |
| PC        | .637 | <u>89.1</u> | .568 | 87.6 | .944 | 83.7 | .728 | 62.7 | .714 | 79.5 | .816 | 69.7 |
| CPC       | .651 | <u>89.1</u> | .592 | 87.6 | .942 | 83.9 | .719 | 62.7 | .715 | 79.3 | .814 | 71.2 |
| PC-Stable | .632 | <u>89.1</u> | .564 | 87.6 | .944 | 85.7 | .725 | 61.1 | .714 | 79.5 | .807 | 67.1 |
| CPC-Stable | .650 | <u>89.1</u> | .548 | 87.6 | .939 | 84.5 | .716 | 62.7 | .738 | 79.3 | .832 | 73.1 |
| GES       | .641 | <u>89.1</u> | .718 | 87.6 | <u>.945</u> | 84.5 | <u>.738</u> | <u>64.2</u> | .743 | 79.3 | .833 | 73.8 |
| IMaGES_BDeu | .641 | <u>89.1</u> | .718 | 87.6 | <u>.945</u> | 84.5 | <u>.738</u> | <u>64.2</u> | .743 | 79.3 | .833 | 73.8 |
| CCD       | .630 | <u>89.1</u> | .563 | 87.6 | .940 | 83.2 | .722 | <u>64.2</u> | .723 | 79.5 | .810 | 71.2 |
| FANG      | .498 | <u>89.1</u> | .447 | 86.3 | .931 | 82.4 | .686 | 59.8 | .499 | 79.3 | .506 | 64.5 |
| FCI       | .633 | <u>89.1</u> | .571 | 87.6 | .944 | 83.7 | .731 | 62.7 | .713 | 79.5 | .816 | 69.7 |
| RFCI      | .649 | <u>89.1</u> | .587 | 87.6 | .943 | 85.7 | .724 | 61.1 | .725 | 79.5 | .811 | 67.1 |
| CFCI      | .592 | <u>89.1</u> | .570 | 87.6 | .922 | 82.1 | .727 | <u>64.2</u> | .762 | 79.3 | .815 | 71.2 |
| GFCI      | .631 | <u>89.1</u> | .696 | 87.6 | <u>.945</u> | 86.0 | .713 | <u>64.2</u> | .680 | 79.5 | .836 | 68.4 |
| FAS       | .682 | <u>89.1</u> | .573 | 87.6 | .939 | 83.2 | .716 | <u>64.2</u> | .767 | 79.3 | .819 | 71.2 |

TABLE 4.

Model statistics when trained with the imputed (complete) forensic dataset, and parameterised with the MLE algorithm. Top performance scores are underlined.

|           | True model ? | | | | Arc comparison stats | | | | |
|-----------|------|------|---|------|----|----|----|----|--------|
| Algorithm | $Chi^2$ | Df | p | BIC | a | d | r | m | DDM |
| KBG       | 4.8k | 562.2k | 1 | -1967k | - | - | - | - | - |
| PC        | 12.6k | 2k | 0 | -42k | 52 | 84 | 16 | 2 | -1.235 |
| CPC       | 13.6k | 1.4k | 0 | -40k | 52 | 84 | 15 | 3 | -1.230 |
| PC-Stable | 12.3k | 1.8k | 0 | -42k | 44 | 88 | 9 | 5 | -1.201 |
| CPC-Stable | 12.6k | 1.1k | 0 | -39k | 44 | 88 | 10 | 4 | -1.206 |
| FGES      | 14.6k | 0.8k | 0 | <u>-37k</u> | 61 | 87 | 5 | 10 | -1.328 |
| IMaGES_BDeu | 14.6k | 0.8k | 0 | <u>-37k</u> | 61 | 87 | 5 | 10 | -1.328 |
| CCD       | 13.5k | 1.5k | 0 | -40k | 51 | 84 | 8 | 10 | -1.186 |
| FANG      | 9.8k | 3.6k | 0 | -49k | 33 | 93 | 8 | 1 | -1.186 |
| FCI       | 12.6k | 2k | 0 | -42k | 52 | 84 | 16 | 2 | -1.235 |
| RFCI      | 11.7k | 2.2k | 0 | -43k | 44 | 89 | 8 | 5 | -1.216 |
| CFCI      | 13.2k | 1.1k | 0 | -39k | 51 | 86 | 9 | 7 | -1.230 |
| GFCI      | 13.1k | 1.7k | 0 | -41k | 56 | 90 | 5 | 7 | -1.338 |
| FAS       | 12.7k | 1.2k | 0 | -39k | 50 | 84 | 5 | 13 | <u>-1.162</u> |

---

[4] We also provide the tables on model statistics under the assumption that missing data are missing at random in Appendix B.

TABLE 5.
Predictive accuracy scores for the primary (P) and secondary (S) variables of interest as specified in the relevant publication [35], based on the models in Table 4. Top performance scores are underlined.

| | Violence (P) | | Anger (S) | | Impulsivity (S) | | Mental illness (S) | | Hazard. drinking (S) | | Substance misuse[5] (S) | |
|---|---|---|---|---|---|---|---|---|---|---|---|---|
| Algorithm | AUC | CC | AUC | CC | AUC | CC | AUC | CC | AUC | CC | AUC | CC |
| KBG | .886 | 82.1 | .624 | 75.7 | .860 | 66.8 | .898 | 83.6 | .946 | 73.9 | .899 | 83.0 |
| PC | .730 | 76.9 | .714 | 76.1 | .779 | 61.3 | .867 | 82.4 | .799 | 70.9 | .915 | 84.6 |
| CPC | .734 | 76.9 | .733 | 76.1 | .778 | 61.3 | .864 | 82.7 | .809 | 70.9 | .928 | 85.5 |
| PC-Stable | .704 | 76.3 | .731 | 76.1 | .780 | 62.2 | .864 | 81.9 | .742 | 70.2 | .918 | 86.3 |
| CPC-Stable | .700 | 76.3 | .725 | 76.1 | .784 | 62.2 | .864 | 82.3 | .771 | 70.2 | .917 | 86.0 |
| FGES | .699 | 76.3 | .751 | 79.3 | .762 | 61.6 | .789 | 81.2 | .839 | 73.6 | .929 | 85.8 |
| IMaGES_BDeu | .699 | 76.3 | .751 | 79.3 | .762 | 61.6 | .789 | 81.2 | .839 | 73.6 | .929 | 85.8 |
| CCD | .736 | 77.3 | .730 | 76.1 | .782 | 60.6 | .858 | 82.5 | .797 | 70.9 | .928 | 86.3 |
| FANG | .702 | 76.3 | .726 | 74.8 | .524 | 46.9 | .662 | 81.2 | .764 | 70.2 | .816 | 79.1 |
| FCI | .735 | 76.9 | .714 | 76.1 | .776 | 61.3 | .867 | 82.4 | .794 | 70.9 | .915 | 84.6 |
| RFCI | .697 | 76.3 | .725 | 76.1 | .763 | 61.6 | .864 | 81.9 | .752 | 70.2 | .913 | 85.8 |
| CFCI | .739 | 77.3 | .682 | 76.1 | .780 | 61.3 | .854 | 82.6 | .802 | 70.9 | .895 | 84.5 |
| GFCI | .677 | 75.67 | .755 | 79.3 | .760 | 61.6 | .821 | 81.3 | .750 | 70.2 | .919 | 85.0 |
| FAS | .746 | 77.3 | .731 | 76.1 | .779 | 60.6 | .854 | 82.4 | .811 | 70.9 | .913 | 84.9 |

TABLE 6.
Model statistics when trained with the imputed (complete) sports dataset, and parameterised with the MLE algorithm. Top performance scores are underlined.

| | True model ? | | | | Arc comparison stats | | | | |
|---|---|---|---|---|---|---|---|---|---|
| Algorithm | $Chi^2$ | $Df$ | $p$ | BIC | a | d | r | m | DDM |
| KBG | 2.8k | 44.9k | 1 | -132.6k | - | - | - | - | - |
| PC | 3.7k | 1.2k | 0 | -7.6k | 8 | 16 | 4 | 2 | -0.909 |
| CPC | 3.7k | 1.1k | 0 | -7.4k | 8 | 16 | 5 | 1 | -0.932 |
| PC-Stable | 3.2k | 0.6k | 0 | -6.1k | 6 | 16 | 4 | 2 | -0.818 |
| CPC-Stable | 3.2k | 0.5k | 0 | -5.8k | 6 | 16 | 6 | 0 | -0.864 |
| PC-Max | 3.5k | 0.4k | 0 | -5.5k | 6 | 16 | 4 | 2 | -0.818 |
| FGES | 4.2k | 0.4k | 0 | -5.2k | 9 | 16 | 2 | 4 | -0.909 |
| IMaGES_BDeu | 4.2k | 0.4k | 0 | -5.2k | 9 | 16 | 2 | 4 | -0.909 |
| CCD | 3.4k | 0.3k | 0 | -5.3k | 8 | 17 | 3 | 2 | -0.977 |
| FANG | 3.2k | 0.3k | 0 | -5.2k | 7 | 16 | 2 | 4 | -0.818 |
| FCI | 3.7k | 1.2k | 0 | -7.6k | 8 | 16 | 1 | 5 | -0.841 |
| RFCI | 3k | 0.6k | 0 | -6.2k | 6 | 16 | 2 | 4 | -0.773 |
| CFCI | 3.3k | 1k | 0 | -7.1k | 8 | 16 | 4 | 2 | -0.909 |
| GFCI | 3.8k | 0.8k | 0 | -6.5k | 7 | 16 | 2 | 4 | -0.818 |
| FAS | 3.2k | 0.4k | 0 | -5.5k | 7 | 16 | 4 | 2 | -0.864 |

TABLE 7.
Predictive accuracy scores for the primary (P) and secondary (S) variables of interest as specified in the relevant publication [36], based on the models in Table 6. Top performance scores are underlined.

| | Next season's league points (P) | | Points dif. from previous season (P) | | Net transfer spending relative to adversaries (S) | | Ability to deal with injuries next season (S) | | EU involvement experience (S) | |
|---|---|---|---|---|---|---|---|---|---|---|
| Algorithm | AUC | CC | AUC | CC | AUC | CC | AUC | CC | AUC | CC |
| KBG | .914 | 55 | .946 | 67 | .873 | 86 | .966 | 87.3 | .929 | 85.3 |
| PC | .711 | 30 | .508 | 25.3 | .909 | 86 | .881 | 78 | .540 | 66 |
| CPC | .723 | 30.3 | .527 | 25.3 | .905 | 86.6 | .886 | 78 | .560 | 66 |
| PC-Stable | .525 | 21 | .483 | 25.3 | .895 | 87.6 | .893 | 78 | .474 | 66 |
| CPC-Stable | .521 | 21 | .496 | 25.3 | .903 | 88 | .897 | 78 | .549 | 66 |
| PC-Max | .464 | 21 | .505 | 25.3 | .904 | 87.3 | .900 | 78 | .565 | 66 |
| FGES | .715 | 30.3 | .523 | 25.3 | .875 | 86 | .947 | 86 | .981 | 88.6 |
| IMaGES_BDeu | .715 | 30.3 | .523 | 25.3 | .875 | 86 | .947 | 86 | .981 | 88.6 |
| CCD | .717 | 30.3 | .523 | 25.3 | .858 | 83.6 | .896 | 78 | .554 | 66 |
| FANG | .511 | 21 | .486 | 25.3 | .856 | 83 | .900 | 78 | .823 | 88.6 |
| FCI | .727 | 30.3 | .533 | 25.3 | .911 | 86 | .896 | 78 | .569 | 66 |
| RFCI | .502 | 21 | .522 | 25.3 | .898 | 87.6 | .902 | 71.6 | .534 | 66 |
| CFCI | .787 | 31.3 | .485 | 25.3 | .887 | 88 | .903 | 71.6 | .560 | 66 |
| GFCI | .793 | 31.3 | .485 | 25.3 | .882 | 86 | .892 | 78 | .809 | 88.6 |
| FAS | .732 | 30.3 | .493 | 25.3 | .870 | 88 | .896 | 78 | .556 | 66 |

TABLE 8.
Model statistics when trained with the first 1,000 samples of the simulated housing market dataset, and parameterised with the MLE algorithm. Top performance scores are underlined.

| | True model ? | | | | Arc comparison stats | | | | |
|---|---|---|---|---|---|---|---|---|---|
| Algorithm | $Chi^2$ | $Df$ | $p$ | BIC | a | d | r | m | DDM |
| KBG | 34k | 3k | 0 | -34.9k | - | - | - | - | - |
| PC | 28k | 2.9k | 0 | -37.1k | 7 | 12 | 9 | 10 | -0.145 |
| CPC | 27k | 23k | 0 | -35.5k | 7 | 12 | 11 | 8 | -0.177 |
| PC-Stable | 26k | 2.9k | 0 | -38.1k | 7 | 12 | 10 | 9 | -0.161 |
| CPC-Stable | 25k | 3.8k | 0 | -42.1k | 7 | 12 | 13 | 6 | -0.210 |
| FGES | 34k | 1.6k | 0 | -29.7k | 4 | 8 | 5 | 18 | 0.274 |
| IMaGES_BDeu | 34k | 1.4k | 0 | -29.3k | 4 | 9 | 4 | 18 | 0.226 |
| CCD | 27k | 3.8k | 0 | -40.8k | 7 | 12 | 2 | 17 | -0.032 |
| FANG | 23k | 9.4k | 0 | -62.2k | 7 | 14 | 9 | 8 | -0.274 |
| FCI | 28k | 2.9k | 0 | -37.1k | 7 | 12 | 5 | 14 | -0.081 |
| RFCI | 28k | 2k | 0 | -34.5k | 7 | 12 | 7 | 12 | -0.113 |
| CFCI | 27k | 2.3k | 0 | -35.5k | 7 | 12 | 8 | 11 | -0.129 |
| GFCI | 31k | 1.6k | 0 | -31.1k | 4 | 10 | 6 | 14 | 0.100 |
| FAS | 26k | 1.4k | 0 | -33k | 7 | 12 | 15 | 4 | -0.242 |

TABLE 9.
Predictive accuracy scores for the primary (P) and secondary (S) variables of interest as specified in the relevant publication [37], based on the models in Table 2. Top performance scores are underlined.

| | Net profit from rental income (P) | | Interest payments (S) | | Borrowing (S) | | Property value (S) | |
|---|---|---|---|---|---|---|---|---|
| Algorithm | AUC | CC | AUC | CC | AUC | CC | AUC | CC |
| KBG | .959 | 74.4 | 1 | 100 | .958 | 79.5 | .991 | 89.5 |
| PC | .787 | 40.4 | 1 | 100 | .928 | 73.7 | .982 | 87.8 |
| CPC | .802 | 40.4 | 1 | 100 | .928 | 73.7 | .982 | 87.8 |
| PC-Stable | .807 | 40.4 | 1 | 100 | .929 | 73.7 | .984 | 87.8 |
| CPC-Stable | .809 | 34.9 | 1 | 100 | .928 | 73.7 | .982 | 87.8 |
| FGES | .959 | 74.4 | 1 | 100 | .955 | 79.9 | .988 | 89.2 |
| IMaGES_BDeu | .959 | 74.4 | 1 | 100 | .955 | 79.9 | .988 | 89.2 |
| CCD | .803 | 40.4 | 1 | 100 | .928 | 73.7 | .985 | 87.7 |
| FANG | .811 | 40.4 | 1 | 100 | .831 | 57.1 | .985 | 87.8 |
| FCI | .804 | 40.4 | 1 | 99.9 | .928 | 73.7 | .982 | 87.8 |
| RFCI | .809 | 40.4 | 1 | 99.9 | .929 | 73.7 | .982 | 87.8 |
| CFCI | .802 | 40.4 | 1 | 100 | .928 | 73.7 | .983 | 87.8 |
| GFCI | .959 | 74.4 | 1 | 100 | .955 | 79.9 | .988 | 89.0 |
| FAS | .804 | 40.4 | 1 | 100 | .825 | 56.5 | .982 | 87.8 |

We have performed 10 tests for each model statistics metric, and seven tests for each predictive accuracy metric. The model statistics are based on two tests for each of the three real-world data case studies, one under the assumption missing data are missing at random and another assuming they are not missing at random (for a total of six tests), and one test for each sample size of synthetic data in case study 4 (for a total of four tests). On the other hand, the predictive accuracy tests consist of one test per case study 1 to 3, restricted to the assumption that missing data are not missing at random (see footnote 3), for a total of three tests, and one test per sample size of synthetic data in case study 4, for a total of four tests. The results are as follows:

 i. **Df:** While *Df* do not tell us much about the accuracy of a model, they can tell us something about the complexity of a model and the extent to which models may be similar. In case studies 1 to 3, KBG generates higher *Df* relative to the algorithms in all but one test, in which FANG generates higher *Df*. In general, all of the algorithms tend to generate similar levels of *Df*, though FANG appears to deviate from them, as well as the KBG, in many of the tests.

[5] This is a summary of three variables; cocaine, cannabis and ecstasy after release.

    In the simulation-based case study 4, *Df* appear to be similar between algorithms and more in line with the KBG when the sample size is relatively low. However, *Df* increase progressively and fluctuate dramatically between algorithms with higher sample size. This is due to the higher number of arcs introduced by the algorithms as the sample size increases, and which we discuss in more detail in (*iv*) below.

 ii. ***p*-value given Chi$^2$:** In case studies 1 to 3, KBG was assessed as a potentially true model in all three cases that assume missing values are not missing at random. When the missing data assumption changes to 'at random', FANG and KBG are assessed as potentially true models in one out of the three case studies (FANG in case study 1 and KBG in case study 2). It is interesting to note that the test in which FANG was assessed as a potentially true model is the same test in which FANG generated the highest *Df* in (*i*) above (refer to Table B.1). None of the residual algorithms is assessed as having produced a potentially true model under any of those six tests. Lastly, none of the algorithms nor KBG is assessed as having produced a true model in any of the four simulation-based tests of case study 4. Overall, KBG and FANG were assessed as a potentially true model in four and one tests respectively, out of a total of 10 tests.

 iii. **BIC:** Fig 3 summarises the results by presenting the percentage of times a model was assessed in the top three[6] (from the tables subsection 3.4 and Appendix C) in terms of BIC score. Since BIC is a model selection measure, it involves 10 tests; the six real-world tests and the four synthetic-based tests. The results show strong inconsistencies between real-world and synthetic-based tests, in terms of which model is best as determined by BIC score.

    KBG is assessed as the worst model, in terms of BIC score, and by a considerable score difference under all six tests in case studies 1 to 3; i.e., all tests that involve learning from real data. This implies that the fitting distributions obtained from the knowledge graphs do not fit the observed distributions as well as those learned from data, and/or that the complexity of the CPTs in KBG is higher compared to those produced by the algorithms. The most likely explanation of this observation is the limited sample size of the real datasets which inevitably guide structure learning towards sparse graphs.

    Interestingly, the BIC performance of KGB in case study 4 is average (or mediocre) in the presence of limited data, as shown in Table 8, and ranks first for higher sample sizes as shown in Appendix C. It is important to clarify that, because synthetic data are generated under the assumption the KBG is the ground truth graph, it is inevitable that KBG will be judged highly on the basis of this assumption and not on the basis of knowledge. The reason the ground truth graph (i.e., KBG) performs mediocre in terms of BIC in Table 8 suggests that maximising BIC in the presence of limited data is unlikely to lead score-based learning towards the search path that contains the ground truth graph, simply because the sample size of the input data is insufficient for reasonably accurate BIC scores to be produced.

    Furthermore, score-based algorithms such as FGES and IMaGES_BDeu whose objective function maximises BIC/BDeu, rank best overall in terms of BIC as expected. Similarly, the hybrid GFCI performed well on synthetic datasets and poorly on real datasets, whereas constraint-based algorithms that do not involve maximising BDeu/BIC naturally fair below score-based and hybrid learning.

 iv. **DDM:** In case studies 1 to 3, all of the algorithms demonstrate a similar pattern in terms of arc comparison to KBG, whereby a very small number of arcs are matched or reversed, a higher number of arcs are introduced, and more than half of KBG's arcs are removed. When the missing data assumption changes from 'missing at random' to 'not missing at random', the DDM scores appear to improve slightly overall. Fig 4 summarises the results in the same way as in Fig 3, and shows that PC-Max[7] and RFCI performed particularly well in terms of DDM score for real data tests, followed by PC-Stable and CPC-Stable.

    In the synthetic-based case study 4, the DDM scores are, overall, considerably higher relative to the DDM scores in real-world data tests, as expected. However, the number of arcs discovered by the algorithms increase aggressively with sample size (refer to Tables 8, C1, C3, and C5), and this behaviour progressively decreases DDM scores. Fig 4 shows that GFCI, FGES, and IMaGES_BDeu, the top three score-based algorithms, performed remarkably well and are the only algorithms that were consistently in the top three in all four synthetic-based tests. It is also worth mentioning that FAS and FANG were consistently inferior relative to the rest of the algorithms, and that PC's DDM score decreased aggressively with the increased sample size, eventually generating the lowest DDM score at 25,000 samples.

---

[6] More than three models may be considered in the case of equivalent performance scores.

[7] Note that we could not get PC-Max to work on the simulation-based study and hence, its overall score in Fig 4 simply mimics its real-world data score.

Finally, the results once more reveal strong inconsistencies between real-world and synthetic data tests. The three algorithms that performed best with synthetic data (GFCI, FGES, and IMaGES_BDeu) performed poorly with real data, and vice versa (for PC-Max, RFCI, PCStable, and CPCStable).

v. **AUC and CC:** Because of the imbalance in the data, greater importance should be given to the AUC statistic over CC. For example, in Table 3 the KBG generates a much higher AUC score for *Violent convictions*, but a lower CC score relative to all of the algorithms. However, only nine violent convictions are recorded in the data over a sample size of 383 instances. Because of this imbalance in the data, a model that always predicts "*no violent convictions*" will have a CC score of 97.65%, but it will be useless in predicting violent convictions.

Figs 5 and 6 illustrate the results in terms of AUC and CC scores respectively, in the same way as in Figs 3 and 4. Unlike previous assessments of model statistics, results from predictive accuracy appear to be fairly consistent between real-world and simulated data, for both AUC and CC scores. Specifically, KBG is assessed as the best performing model in terms of AUC assessments in real data tests since it ranked in the top three in 12 out of the 17 (70.59%) variables of interests (refer to the Tables 3, 5, and 7), followed by FGES and IMaGES_BDeu with 52.94% frequency in top 3. For the synthetic-based case study 4, FGES ranks best with 81.25%, IMaGES_BDeu second with 75%, and KBG third with 62.5%; this is despite KBG representing the ground truth graph, and this observation suggests some graphical models may have overfitted the data, and supports the notion that predictive accuracy is a poor measure of causality.

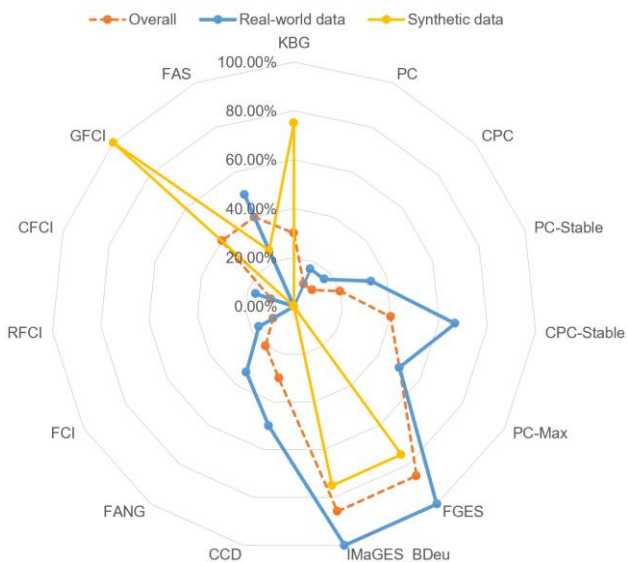

**Fig 3.** The percentage of times a constructed model ranked in the top three in terms of BIC score.

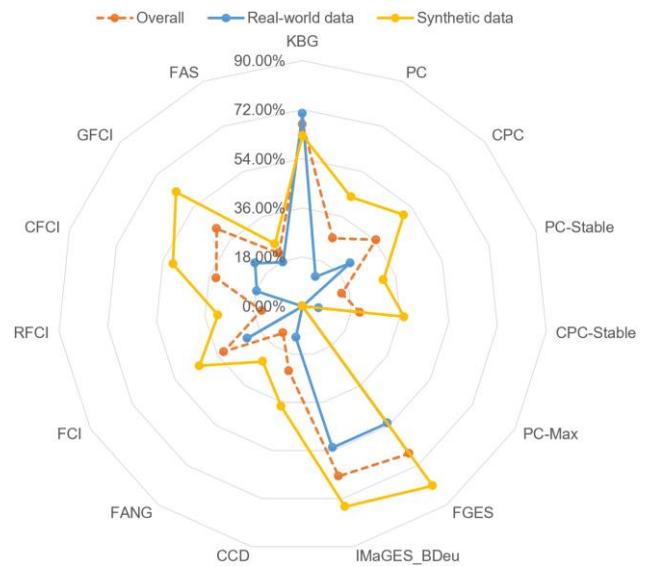

**Fig 5.** The percentage of times a constructed model ranked in the top three in terms of AUC score.

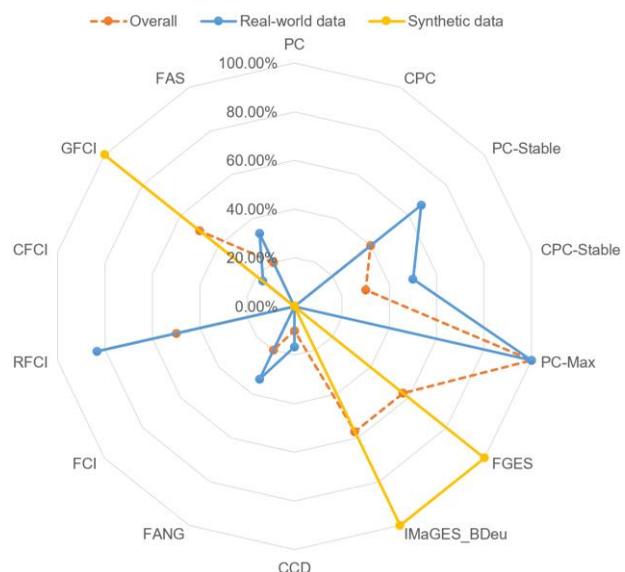

**Fig 4.** The percentage of times a constructed model ranked in the top three in terms of DDM score.

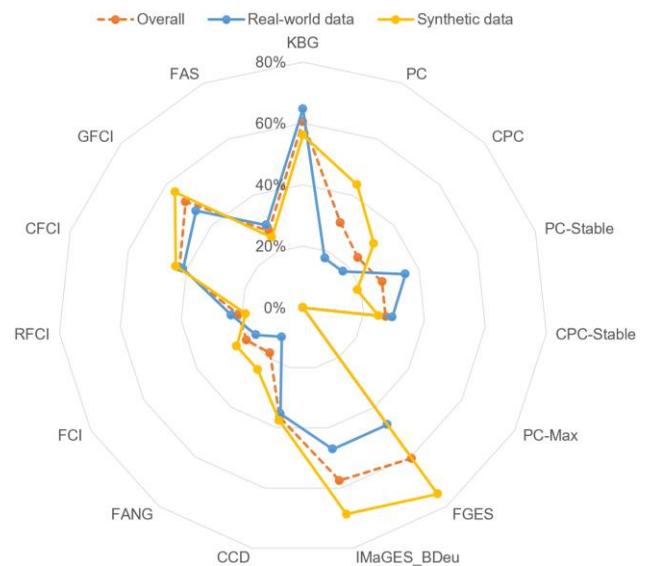

**Fig 6.** The percentage of times a constructed model ranked in the top three in terms of CC score.

Regardless of some imbalances in the data, the CC results tend to mirror the AUC results, at least when it comes to the top three performers. It is also important to note that when KBG performs best in terms of AUC score, the AUC difference from the machine learned models is often major (especially when it comes to real-world data tests), whereas when a machine learned model performs best the difference relative to the KBG's performance tends to be considerably smaller.

The study inevitably suffers from some limitations. The limitations arise either due to the software used to test the different algorithms, the algorithms themselves, the available data, the methodology, or any combination of the above. The key limitations are:

i. As previously mentioned, this study assumes that both the machine learned and knowledge-based graphs are restricted to the same set of measured variables and hence, we do not consider knowledge-based latent variables which had initially been introduced in the case studies. This restriction may bias the results against the knowledge-based models under the assumption that the knowledge-based latent variables are typically introduced as known causal and explanatory variables that are important for causal inference but missing from the data. On the other hand, an argument can be made that most of these algorithms assume causal sufficiency and are not suitable for such data. However, such an argument would make all algorithms unsuitable for real data analysis, and this is because real data incorporate various forms of noise that violate the data assumptions of most algorithms.

ii. The algorithms tested are restricted to those available in TETRAD, some of which are different variants of the same algorithm; e.g., The CPC, PC-Stable, CPC-Stable and PC-Max are variants of PC. We found that extending the assessment of the algorithms across different structure learning software is impractical since not all of the statistical measures available in TETRAD are available in other software, and those that are available are often defined differently, thereby not enabling direct comparisons. While TETRAD covers a large range of different algorithms, the vast majority of them are constraint-based; although we have included algorithms spanning all three classes of learning.

iii. Because it is impractical to test how each algorithm behaves with different parameter settings, but also because this is out of the scope of this paper, the algorithms are assessed based on their default input parameters as implemented in TETRAD.

iv. For the reasons discussed in Section 2.2, we have restricted our assessment to discrete variables. As a result, algorithms that can only work with continuous data have not been considered (see Table A.1).

v. When an algorithm preserves undirected or bidirected edges, it sometimes does so under the assumption that the directionality of the edge does not produce a different model. Other times, however, an algorithm may simply fail to determine the directionality of an edge. Irrespective of the reason, and because we are interested in CBNs, we used a relevant feature in TETRAD to direct undirected or bidirected edges. Since the output of some of these algorithms is a Completed Partially Directed Acyclic Graph (CPDAG) that represents a set of Markov equivalent DAGs, this step obtains one DAG from the equivalence class of DAGs. Otherwise, the edges are directed at random and, because an undirected edge does not imply that each possible direction has equal probability of occurring, it is possible that such randomisation impedes the performance of the algorithms.

vi. Predictive evaluation measures to how well the fitting distributions agree with the empirical distributions. TETRAD does not offer cross-validation, which means that some of the accuracy scores, in terms of CC or AUC as described in Section 2.4, may be subject to model overfitting.

vii. Further to (v) above, the results do not take into consideration how some algorithms may benefit from knowledge-based constraints, but also do not consider algorithms that go beyond the option of introducing knowledge-based constraints and make it a requirement. For instance, Verma and Pearl's IC algorithm [9] depends upon knowledge of the actual conditional independencies between variables, and Cooper and Herskovits' K2 algorithm [53] requires that temporal order is provided in ordered tiers.

viii. There are more algorithms implemented in TETRAD that could have been assessed in this study, but which we could not include because of unknown runtime errors (see Table A.1).

## 4 CONCLUDING REMARKS

The paper provided a direct performance comparison between a set of previously constructed knowledge-based BNs and machine learned BNs. As expected, the algorithms are better at arriving at a graph with a high model selection score, such as BIC. However, this result does not translate into higher predictive accuracy. The results show that many of the highest model selection BNs are weaker predictors of variables of interest, relative to knowledge-based BNs with lower model selection scores.

Despite promising developments with constraint-based and score-based algorithms, it is fair to say that conditional independence tests and maximisation of the given objective functions appear to be inadequate for BN structure discovery when the input data are limited in terms of sample size. Because many real-world problems involve high dimensionality datasets with relatively limited sample size, a greater focus should be placed on BN structure learning to better address these problems and achieve the kind of automated learning that is believed to be necessary for causal discovery with real data. This challenge highlights the value of causal knowledge, since the two approaches to causal structure learning can indeed be complementary; i.e., machine learned graphs can be refined by domain experts who have the ability to leverage substantive knowledge that may not be well-represented in the data.

The results support the views of [14] on the negative repercussions from not having an agreed evaluation process to assess the usefulness of these algorithms. The different scoring metrics considered in this paper do confirm such inconsistencies, and this is problematic since, in the absence of an agreed evaluation method it is difficult to reach a consensus about a) whether a particular algorithm is sufficiently accurate, and b) which of the competing algorithms is superior. However, it should be noted that this is a wider issue not limited to this field; e.g., in economics researchers have shown that there is a weak relationship between different summary error statistics and profit measures [54], [55].

Finally, the results also reveal that it is possible for algorithms that demonstrate stronger performance on real-world data to demonstrate weaker performance on simulated data, and vice versa (refer to Figs 3 and 4). Hence, we found that conclusions about which algorithm is superior to be highly contradictory between real-world and synthetic-based assessments. These results support the views of [8], [27], [56], [57], who argued that synthetic data tell us very little about the extent to which modelling assumptions hold true for real-world applications.

# APPENDIX A

TABLE A.1.
TETRAD algorithms excluded from the assessment process.

| Reason for exclusion | Algorithms excluded |
|---|---|
| Cannot handle missing values[8] | FGES, IMaGES_BDeu, IMaGES_SEM_BIC, IMaGES_CCD, GFCI, TsGFCI. |
| Dataset must be continuous | IMaGES_SEM_BIC, IMaGES_CCD, FgesMb, MGM, GLASSO, Bpc, Fofc, Ftfc. |
| Software unknown error | PC-Max[9], CCD_MAX, TsFCI, TsGFCI, TsImages. |
| Other known error/restriction | MBFS, EB, R1, R2, R3, R4, RSkew, RSkewE, Skew, SkewE, Tahn. |

# APPENDIX B: MODEL STATISTICS FOR CASE STUDIES 1, 2, AND 3 WHEN ASSUMING DATA ARE MISSING AT RANDOM

TABLE B.1.
Model statistics when trained with the medical dataset from case study 1, which incorporates missing values, and parameterised with the EM algorithm. Top performance scores are underlined.

| | True model ? | | | | Arc comparison stats | | | | |
|---|---|---|---|---|---|---|---|---|---|
| Algorithm | $Chi^2$ | $Df$ | $p$ | BIC | $a$ | $d$ | $r$ | $m$ | DDM |
| KBG | 6.3k | 3.3k | 0 | -18.2k | - | - | - | - | - |
| PC | 3.8k | 0.2k | 0 | -10.4k | 26 | 74 | 5 | 5 | -1.101 |
| CPC | 4.4k | 0.3k | 0 | -10.2k | 27 | 74 | 4 | 6 | -1.107 |
| PC-Stable | 3.6k | 0.1k | 0 | -10k | 17 | 75 | 6 | 3 | -1.024 |
| CPC-Stable | 3.7k | 0.1k | 0 | -10k | 17 | 75 | 4 | 5 | -1.012 |
| CCD | 4.4k | 0.2k | 0 | -10k | 27 | 74 | 7 | 3 | -1.125 |
| FANG | 2.6k | 23.4k | 1 | <u>-8k</u> | 28 | 79 | 4 | 1 | -1.238 |
| FCI | 3.8k | 0.2k | 0 | -10.4k | 27 | 74 | 5 | 5 | -1.113 |
| RFCI | 3k | 0.1k | 0 | -10.5k | 16 | 75 | 4 | 5 | <u>-1.000</u> |
| CFCI | 3.8k | 0.2k | 0 | -10.2k | 27 | 74 | 4 | 6 | -1.107 |
| FAS | 3.9k | 0.1k | 0 | -10k | 27 | 75 | 4 | 5 | -1.131 |

TABLE B.2.
Model statistics when trained with the forensic dataset from case study 2, which incorporates missing values, and parameterised with the EM algorithm. Top performance scores are underlined.

| | True model ? | | | | Arc comparison stats | | | | |
|---|---|---|---|---|---|---|---|---|---|
| Algorithm | $Chi^2$ | $Df$ | $p$ | BIC | $a$ | $d$ | $r$ | $m$ | DDM |
| KBG | 13.4k | 9.1k | 0 | -60k | - | - | - | - | - |
| PC | 13.3k | 0.4k | 0 | -31k | 48 | 88 | 11 | 4 | -1.228 |
| CPC | 12.5k | 0.3k | 0 | -31k | 48 | 88 | 10 | 5 | -1.223 |
| PC-Stable | 11.7k | 0.4k | 0 | -32k | 38 | 92 | 9 | 2 | -1.199 |
| CPC-Stable | 11.5k | 0.3k | 0 | -31k | 38 | 92 | 7 | 4 | -1.189 |
| CCD | 12.6k | 0.4k | 0 | -31k | 47 | 88 | 6 | 9 | -1.194 |
| FANG | 5.6k | 0.4k | 0 | -35k | 43 | 96 | 5 | 2 | -1.306 |
| FCI | 12.3k | 0.4k | 0 | -31k | 48 | 88 | 11 | 4 | -1.228 |
| RFCI | 10.7k | 0.4k | 0 | -32k | 38 | 92 | 6 | 5 | <u>-1.184</u> |
| CFCI | 10.8k | 0.3k | 0 | -32k | 47 | 88 | 8 | 7 | -1.204 |
| FAS | 13k | 0.2k | 0 | <u>-30k</u> | 44 | 89 | 6 | 8 | <u>-1.184</u> |

TABLE B.3.
Model statistics when trained with the sports dataset from case study 3, which incorporates missing values, and parameterised with the EM algorithm. Top performance scores are underlined.

| | True model ? | | | | Arc comparison stats | | | | |
|---|---|---|---|---|---|---|---|---|---|
| Algorithm | $Chi^2$ | $Df$ | $p$ | BIC | $a$ | $d$ | $r$ | $m$ | DDM |
| KBG | 3.8k | 12.6k | 1 | -38.9k | - | - | - | - | - |
| PC | 2.9k | 0.8k | 0 | -5.6k | 7 | 17 | 4 | 1 | -0.955 |
| CPC | 2.8k | 0.6k | 0 | -5.1k | 7 | 17 | 5 | 0 | -0.977 |
| PC-Stable | 2.3k | 0.2k | 0 | <u>-4.3k</u> | 4 | 18 | 2 | 2 | <u>-0.864</u> |
| CPC-Stable | 2.1k | 0.2k | 0 | <u>-4.3k</u> | 4 | 18 | 4 | 0 | -0.909 |
| PC-Max | 2.3k | 0.2k | 0 | <u>-4.3k</u> | 4 | 18 | 2 | 2 | <u>-0.864</u> |
| CCD | 3.3k | 0.3k | 0 | <u>-4.3k</u> | 7 | 17 | 1 | 4 | -0.886 |
| FANG | 2.6k | 0.4k | 0 | -4.8k | 12 | 16 | 3 | 3 | -1.068 |
| FCI | 2.9k | 0.8k | 0 | -5.6k | 7 | 17 | 2 | 3 | -0.909 |
| RFCI | 2k | 0.2k | 1 | -4.5k | 4 | 18 | 2 | 2 | <u>-0.864</u> |
| CFCI | 2.8k | 0.6k | 0 | -5.2k | 7 | 17 | 2 | 3 | -0.909 |

---

[8] Note that some of the algorithms reporting this restriction were later assessed with the complete datasets.

[9] This algorithm would work for some, but not all, of our case studies.

# APPENDIX C: MODEL STATISTICS FOR CASE STUDY 4 WHEN BASED ON HIGHER SAMPLE SIZES

TABLE C.1.

Model statistics when trained with the first 5,000 samples of the simulated housing market dataset, and parameterised with the MLE algorithm. Top performance scores are underlined.

| Algorithm | True model? | | | | Arc comparison stats | | | | |
|---|---|---|---|---|---|---|---|---|---|
|  | $Chi^2$ | $Df$ | $p$ | BIC | $a$ | $d$ | $r$ | $m$ | DDM |
| KBG | 173k | 3k | 0 | <u>-132k</u> | - | - | - | - | - |
| PC | 163k | 16k | 0 | -191k | 18 | 7 | 7 | 17 | -0.145 |
| CPC | 170k | 24k | 0 | -224k | 18 | 7 | 9 | 15 | -0.177 |
| PC-Stable | 150k | 44k | 0 | -316k | 15 | 8 | 12 | 11 | -0.194 |
| CPC-Stable | 155k | 14k | 0 | -189k | 15 | 8 | 11 | 12 | -0.177 |
| FGES | 178k | 6k | 0 | -143k | 13 | 5 | 8 | 18 | <u>0.129</u> |
| IMaGES_BDeu | 178k | 6k | 0 | -143k | 13 | 5 | 8 | 18 | <u>0.129</u> |
| CCD | 165k | 10k | 0 | -166k | 17 | 9 | 5 | 17 | -0.210 |
| FANG | 137k | 234k | 0 | -1135k | 20 | 10 | 11 | 10 | -0.468 |
| FCI | 163k | 15k | 0 | -186k | 18 | 7 | 7 | 17 | -0.145 |
| RFCI | 145k | 35k | 0 | -280k | 15 | 9 | 7 | 15 | -0.177 |
| CFCI | 171k | 22k | 0 | -214k | 18 | 7 | 4 | 20 | -0.097 |
| GFCI | 162k | 2k | 0 | -133k | 10 | 11 | 3 | 17 | -0.081 |
| FAS | 139k | 6k | 0 | -161k | 18 | 7 | 19 | 5 | -0.339 |

TABLE C.2.

Predictive accuracy scores for the primary (P) and secondary (S) variables of interest as specified in the relevant publication [37], based on the models in Table C1. Top performance scores are underlined.

| Algorithm | Net profit from rental income (P) | | Interest payments (S) | | Borrowing (S) | | Property value (S) | |
|---|---|---|---|---|---|---|---|---|
|  | AUC | CC | AUC | CC | AUC | CC | AUC | CC |
| KBG | <u>.962</u> | <u>74.7</u> | 1 | 100 | .953 | 78.2 | <u>.992</u> | <u>90.6</u> |
| PC | .903 | 54.4 | 1 | 100 | .952 | 78 | .989 | 88.4 |
| CPC | .903 | 54.4 | 1 | 100 | .952 | 78 | .991 | 90.1 |
| PC-Stable | .807 | 40 | 1 | 100 | .940 | 78 | .986 | 88.6 |
| CPC-Stable | .772 | 29 | 1 | 100 | .952 | 78 | .989 | 88.4 |
| FGES | <u>.962</u> | <u>74.7</u> | 1 | 100 | <u>.954</u> | <u>78.7</u> | <u>.992</u> | 90.4 |
| IMaGES_BDeu | <u>.962</u> | <u>74.7</u> | 1 | 100 | <u>.954</u> | <u>78.7</u> | <u>.992</u> | 90.4 |
| CCD | .877 | 49.9 | 1 | 100 | .929 | 73.2 | .991 | 90 |
| FANG | .866 | 44.8 | 1 | 100 | .821 | 55.9 | .987 | 88.5 |
| FCI | .903 | 54.4 | 1 | 100 | .952 | 78 | .989 | 88.4 |
| RFCI | .808 | 40 | 1 | 100 | .952 | 78 | .986 | 88.2 |
| CFCI | .903 | 54.4 | 1 | 100 | .952 | 78 | .992 | <u>90.6</u> |
| GFCI | <u>.962</u> | <u>74.7</u> | 1 | 100 | .951 | 78 | .984 | 88.5 |
| FAS | .877 | 49.9 | 1 | 100 | .821 | 55.9 | .988 | 88.6 |

TABLE C.3.

Model statistics when trained with the first 10,000 samples of the simulated housing market dataset, and parameterised with the MLE algorithm. Top performance scores are underlined.

| Algorithm | True model? | | | | Arc comparison stats | | | | |
|---|---|---|---|---|---|---|---|---|---|
|  | $Chi^2$ | $Df$ | $p$ | BIC | $a$ | $d$ | $r$ | $m$ | DDM |
| KBG | 347k | 3k | 0 | <u>-250k</u> | - | - | - | - | - |
| PC | 314K | 126k | 0 | -834k | 22 | 5 | 9 | 17 | -0.177 |
| CPC | 353k | 71k | 0 | -562k | 22 | 5 | 7 | 19 | -0.145 |
| PC-Stable | 329k | 31k | 0 | -389k | 20 | 4 | 6 | 21 | 0.000 |
| CPC-Stable | 352k | 50k | 0 | -463k | 20 | 4 | 7 | 20 | -0.016 |
| FGES | 361k | 23k | 0 | -335k | 19 | 4 | 4 | 23 | 0.065 |
| IMaGES_BDeu | 360k | 23k | 0 | -335k | 18 | 4 | 5 | 22 | 0.081 |
| CCD | 344k | 183k | 0 | -1081k | 22 | 6 | 3 | 22 | -0.145 |
| FANG | 288k | 109k | 0 | -767k | 27 | 7 | 10 | 14 | -0.484 |
| FCI | 314k | 126k | 0 | -832k | 22 | 5 | 5 | 21 | -0.113 |
| RFCI | 320k | 94k | 0 | -681k | 20 | 4 | 7 | 20 | -0.016 |
| CFCI | 350k | 64k | 0 | -528k | 22 | 5 | 4 | 22 | -0.097 |
| GFCI | 331k | 5k | 0 | -269k | 12 | 7 | 2 | 22 | <u>0.129</u> |
| FAS | 291k | 31k | 0 | -410k | 22 | 5 | 14 | 12 | -0.258 |

TABLE C.4.

Predictive accuracy scores for the primary (P) and secondary (S) variables of interest as specified in the relevant publication [37], based on the models in Table C4. Top performance scores are underlined.

| Algorithm | Net profit from rental income (P) | | Interest payments (S) | | Borrowing (S) | | Property value (S) | |
|---|---|---|---|---|---|---|---|---|
|  | AUC | CC | AUC | CC | AUC | CC | AUC | CC |
| KBG | .963 | 74.7 | 1 | <u>100</u> | .954 | 78.7 | .992 | 90.5 |
| PC | .968 | 75 | 1 | <u>100</u> | <u>.962</u> | <u>80.9</u> | .992 | 90.3 |
| CPC | <u>.970</u> | 76.6 | 1 | 99.99 | .961 | 80.8 | <u>.993</u> | <u>91.1</u> |
| PC-Stable | .964 | 74.8 | 1 | 99.99 | .935 | 74 | .991 | 89.7 |
| CPC-Stable | .964 | 74.8 | 1 | 99.99 | .958 | 79.4 | <u>.993</u> | <u>91.1</u> |
| FGES | .962 | 74.7 | 1 | <u>100</u> | .955 | 78.9 | <u>.993</u> | <u>91.1</u> |
| IMaGES_BDeu | .962 | 74.7 | 1 | <u>100</u> | .955 | 78.9 | .992 | 90.7 |
| CCD | <u>.970</u> | 76.1 | .997 | 94.27 | .956 | 80.3 | <u>.993</u> | <u>91.1</u> |
| FANG | .833 | 42.5 | 1 | <u>100</u> | .819 | 55.3 | .989 | 89.7 |
| FCI | .968 | 75 | 1 | <u>100</u> | <u>.962</u> | <u>80.9</u> | .991 | 90 |
| RFCI | .965 | 75.1 | 1 | <u>100</u> | .910 | 68.9 | .991 | 89.8 |
| CFCI | <u>.970</u> | 76.7 | 1 | <u>100</u> | .961 | 80.8 | <u>.993</u> | <u>91.1</u> |
| GFCI | .962 | 74.7 | 1 | <u>100</u> | .955 | 78.8 | .991 | 89.9 |
| FAS | .926 | 64.6 | 1 | <u>100</u> | .909 | 68.2 | .989 | 89.4 |

TABLE C.5.

Model statistics when trained with the first 25,000 samples of the simulated housing market dataset, and parameterised with the MLE algorithm. Top performance scores are underlined.

| Algorithm | True model? | | | | Arc comparison stats | | | | |
|---|---|---|---|---|---|---|---|---|---|
|  | $Chi^2$ | $Df$ | $p$ | BIC | $a$ | $d$ | $r$ | $m$ | DDM |
| KBG | 820k | 3k | 0 | <u>-630k</u> | - | - | - | - | - |
| PC | 859k | 155k | 0 | -1380k | 29 | 3 | 7 | 17 | -0.426 |
| CPC | 882k | 166k | 0 | -1423k | 29 | 3 | 6 | 19 | -0.357 |
| PC-Stable | 884k | 199k | 0 | -1590k | 28 | 3 | 3 | 21 | -0.315 |
| CPC-Stable | 881k | 153k | 0 | -1359k | 28 | 3 | 6 | 20 | -0.276 |
| FGES | 917k | 52k | 0 | -831k | 21 | 2 | 6 | 23 | 0.097 |
| IMaGES_BDeu | 917k | 52k | 0 | -831k | 21 | 2 | 6 | 22 | 0.067 |
| CCD | 826k | 410k | 0 | -2688k | 28 | 5 | 2 | 22 | -0.345 |
| FANG | 737k | 1744k | 0 | -9486k | 32 | 3 | 17 | 14 | -0.368 |
| FCI | 859k | 154k | 0 | -1374k | 29 | 3 | 4 | 21 | -0.321 |
| RFCI | 859k | 264k | 0 | -1933k | 28 | 3 | 3 | 20 | -0.365 |
| CFCI | 847k | 64k | 0 | -926k | 29 | 3 | 8 | 22 | -0.182 |
| GFCI | 850k | 24k | 0 | -721k | 16 | 4 | 5 | 22 | <u>0.145</u> |
| FAS | 769k | 13k | 0 | -708k | 29 | 3 | 15 | 12 | -0.417 |

TABLE C.6.

Predictive accuracy scores for the primary (P) and secondary (S) variables of interest as specified in the relevant publication [37], based on the models in Table C5. Top performance scores are underlined.

| Algorithm | Net profit from rental income (P) | | Interest payments (S) | | Borrowing (S) | | Property value (S) | |
|---|---|---|---|---|---|---|---|---|
|  | AUC | CC | AUC | CC | AUC | CC | AUC | CC |
| KBG | .962 | 74.6 | 1 | <u>100</u> | .956 | 78.5 | .990 | 88.6 |
| PC | .926 | 58.2 | 1 | <u>100</u> | <u>.960</u> | 79.6 | .993 | 91.3 |
| CPC | .966 | 75.4 | 1 | 99.996 | .958 | 79.1 | .993 | <u>91.4</u> |
| PC-Stable | .969 | 75.8 | .999 | 98.54 | <u>.960</u> | 79.5 | .993 | 91.1 |
| CPC-Stable | .966 | 75.4 | 1 | 99.996 | .958 | 79.1 | .993 | <u>91.4</u> |
| FGES | <u>.975</u> | <u>79.2</u> | 1 | <u>100</u> | .958 | 79.1 | .993 | 91.2 |
| IMaGES_BDeu | <u>.975</u> | <u>79.2</u> | 1 | <u>100</u> | .958 | 79.1 | .993 | 91.2 |
| CCD | .874 | 49.6 | 1 | <u>100</u> | .958 | 79.2 | .993 | 91.3 |
| FANG | .951 | 70.1 | <u>1</u> | <u>100</u> | .818 | 55.1 | .989 | 89.4 |
| FCI | .926 | 58.2 | <u>1</u> | 99.74 | <u>.960</u> | 79.6 | .993 | 90.9 |
| RFCI | .968 | 75.8 | .999 | 98.54 | <u>.960</u> | 79.5 | .993 | 91.1 |
| CFCI | .966 | 75.1 | <u>1</u> | 99.99 | .958 | 79.1 | <u>.994</u> | 90.8 |
| GFCI | <u>.975</u> | <u>79.2</u> | <u>1</u> | <u>100</u> | .958 | 79.1 | .992 | 90.4 |
| FAS | .964 | 74.8 | <u>1</u> | <u>100</u> | .818 | 55.1 | .989 | 89.3 |

## ACKNOWLEDGMENTS

This work was supported by the ERSRC project EP/S001646/1 on *Bayesian Artificial Intelligence for Decision Making under Uncertainty* [58], by the European Research Council (ERC) project ERC-2013-AdG339182 (BAYES_KNOWLEDGE), and by the EPSRC project EP/P009964/1: PAMBAYESIAN: Patient Managed decision-support using Bayes Networks.

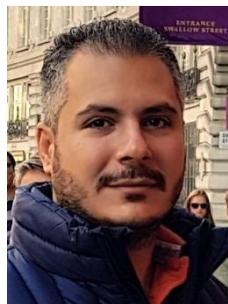

**Anthony C. Constantinou** is a Lecturer (Assistant Prof) in Machine Learning and Data Mining at Queen Mary University of London. His research interests are in Bayesian Artificial Intelligence for causal discovery and intelligent decision making under uncertainty.

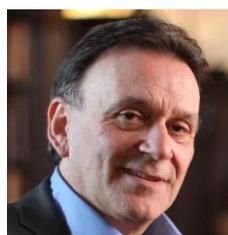

**Norman Fenton** is Professor of Risk Information Management at Queen Mary, University of London and is also a Director of Agena, a company that develops Bayesian probabilistic reasoning software and applies it to risky and uncertain problems.

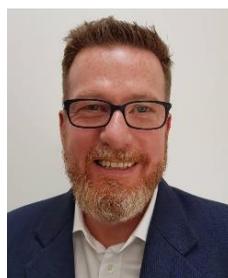

**Martin Neil** is Professor of Computer Science and Statistics at Queen Mary, University of London and is also a Director of Agena, a company that develops Bayesian probabilistic reasoning software and applies it to risky and uncertain problems.